\documentclass[10pt,twocolumn,letterpaper]{article}

\usepackage{multirow}
\usepackage{graphicx}
\usepackage{pifont}
\usepackage{paralist}
\usepackage[accsupp]{axessibility} 
\usepackage[pagenumbers]{iccv} 

\definecolor{iccvblue}{rgb}{0.21,0.49,0.74}
\usepackage{xcolor}
\usepackage{colortbl}
\usepackage[pagebackref,breaklinks,colorlinks,allcolors=iccvblue]{hyperref}

\def\paperID{11609} 
\def\confName{ICCV}
\def\confYear{2025}

\title{Taming the Untamed: Graph-Based Knowledge Retrieval and Reasoning \\ for MLLMs to Conquer the Unknown}

\author{Bowen Wang$^{1}$, Zhouqiang Jiang$^{2}$, Yasuaki Susumu$^{3}$,  Shotaro Miwa$^{3}$, Tianwei Chen$^{2}$, Yuta Nakashima$^{4}$\\
Osaka University, Japan\\
Mitsubishi Electric Corp., Japan\\
{\tt\small $^1$wang@ids.osaka-u.ac.jp}      {\tt\small $^2$\{zhouqiang, chentw\}@is.ids.osaka-u.ac.jp}\\
{\tt\small $^3$\{susumu.yasuaki@bx, miwa.shotaro@bc\}.mitsubishielectric.co.jp}\\
{\tt\small $^4$n-yuta@im.sanken.osaka-u.ac.jp}\\
}

\begin{document}
\maketitle
\begin{abstract}
The real value of knowledge lies not just in its accumulation, but in its potential to be harnessed effectively to conquer the unknown. Although recent multimodal large language models (MLLMs) exhibit impressing multimodal capabilities, they often fail in rarely encountered domain-specific tasks due to limited relevant knowledge. To explore this, we adopt visual game cognition as a testbed and select ``Monster Hunter: World'' as the target to construct a multimodal knowledge graph (MH-MMKG), which incorporates multi-modalities and intricate entity relations. We also design a series of challenging queries based on MH-MMKG to evaluate the models’ ability for complex knowledge retrieval and reasoning. Furthermore, we propose a multi-agent retriever that enables a model to autonomously search relevant knowledge without additional training. Experimental results show that our approach significantly enhances the performance of MLLMs, providing a new perspective on multimodal knowledge-augmented reasoning and laying a solid foundation for future research.\footnote{The dataset and code at https://github.com/wbw520/MH-MMKG.}
\end{abstract}    
\section{Introduction}
\label{sec:intro}
Recent multimodal large language models (MLLMs) \cite{yin2023survey}, particularly closed-source ones \cite{achiam2023gpt,TheC3}, have demonstrated human-like multimodal capabilities, achieving outstanding performance on benchmarks related to commonsense \cite{schwenk2022okvqa}, scientific facts \cite{yue2024mmmu}, \etc. Meanwhile, for domain-specific tasks or rarely seen data \cite{cui2024survey,chen2024can}, relying solely on their own perception and built-in knowledge is inadequate for predicting accurate answers and is hardly interpretable \cite{wang2024exploring,Guan_2024_CVPR}.

To alleviate these issues, multimodal RAG (mRAG) \cite{zhao2024retrieval,zhao2023retrieving} has been introduced to improve MLLM performance through heuristic \cite{gui2022kat,chen2022murag,bonomo2025visual} or agent-based methods \cite{su2024hybrid,li2024benchmarking}. However, as with RAG for LLMs \cite{gao2023retrieval}, such knowledge retrieval approaches often suffer from knowledge redundancy and low relevance \cite{peng2024graph}. 
As a result, knowledge graph retrieval \cite{jin2024large,edge2024local} has gained increasing attention. Knowledge graphs (KGs) \cite{ji2021survey}, storing knowledge in a structured and comprehensive manner, offer models with more context-rich information. As a substantial multimodal extension, multimodal KGs (MMKGs) \cite{chen2024knowledge,lymperaiou2024survey} can offer a knowledge foundation for enhancing MLLMs.

\begin{figure}[t]
    \centering
    \includegraphics[width=1\columnwidth]{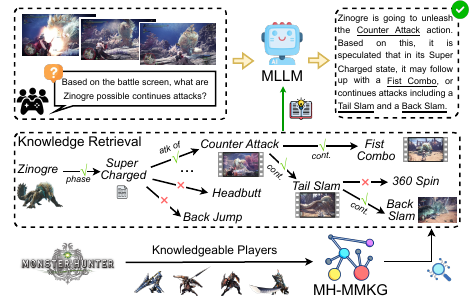}
    \caption{Our MH-MMKG is curated by knowledgeable players. By leveraging a multi-agents retriever, the MLLM's responses can be augmented.}
    \label{fig:fig1}
    \vspace{-0.1in}
\end{figure}

One major challenge MLLMs with MMKGs is the unavailability of robust benchmarks. Existing work primarily relies on VQA datasets \cite{krishna2017visual} or web resources \cite{bollacker2008freebase,ferrada2017imgpedia} for constructing MMKGs. It is highly plausible that such knowledge has already been learned by current MLLMs, making them less useful in evaluating the effectiveness of MMKGs. Additionally, existing MMKGs primarily incorporate text and images, lacking more information-rich modalities, such as video \cite{nguyen2024video}, which reduces their suitability for increasingly complex multimodal tasks. 

Another challenge is how to retrieve knowledge accurately. The mainstream approach embeds entities in MMKGs to find relevant subgraphs \cite{ishiwatari2020relation,guo2021multi,li2023multi,lee-etal-2024-multimodal}. However, training an effective retriever is data-intensive and unrealistic for low-resource domains. In addition, incorporating graphs about rarely seen domains into powerful closed-source models remains impractical. 

This work aims to endeavor the ability of MLLMs to tackle domain-specific tasks by leveraging well-structured external knowledge sources. To achieve this, we build a testbed with a well-known game title \cite{chen2024can} as illustrated in Figure \ref{fig:fig1} for two reasons: First, the visual modality exhibits a large gap from real-world scenes as its visual world is computer graphics-generated mostly with imaginary cultures, creatures, \etc to which MLLMs' perception module is not well exposed. Second, the knowledge of the game's world can be different from the real world, which makes the MLLMs' built-in knowledge less useful. We in this work use ``Monster Hunter: World'' for our testbed as the game series offers abundant knowledge about the fantasy world.\footnote{As one of the most popular game titles, current MLLMs have already learned some knowledge about it. We will experimentally show its impact.} 

To this end, we construct MH-MMKG, which integrates text, images, videos, and intricate entity relations curated by knowledgeable players. Additionally, we design 238 carefully crafted question-answer pairs as benchmark that cover various sub-tasks, including fine-grained visual cognition, conditional reasoning, \etc. Importantly, the knowledge required to answer these questions is encompassed in MH-MMKG. An MLLM can answer most questions with perfect knowledge retrieval and comprehension. MH-MMKG, as well as the benchmark, offers a new dimension of challenges for the community: MH-MMKG requires flexible perception to comprehend the visual modality in an (almost) unseen domain as well as reasoning without relying on the built-in knowledge of the MLLM.

Multi-modal knowledge retrieval through graph is the foundation for tackling this new set of challenges. We thus develop a multi-agent method, which harnesses the self-searching capabilities \cite{jiang2023structgpt,sunthink,fang2024karpa} of MLLMs, allowing them to autonomously retrieve relevant knowledge without training. On both close- and open-source leading MLLMs, we experimentally show the effectiveness of our method, qualifying it as a strong baseline.

Our contribution lies in developing a high-quality benchmark and exploring MLLMs' ability to find knowledge from MMKG for solving rarely seen domain-specific tasks. These tasks require not only advanced multimodal perception but also a deep understanding of complex dependencies, conditions, and rules within MH-MMKG. Our benchmark, together with the baseline, provides a strong foundation for advancing MLLMs toward real-world challenges.
\section{Related Works}
\label{sec:related}

\subsection{Multimodal Retrieval Augmented Generation}
Recent literature shows a growing surge of interest in MLLMs \cite{yin2023survey}. Despite their advancements, even sophisticated models like GPT-4o face difficulty in handling domain-specific tasks \cite{cui2024survey,chen2024can} or reasoning \cite{wang2024exploring,Guan_2024_CVPR}. To mitigate these issues, mRAG \cite{zhao2024retrieval,zhao2023retrieving} seeks to enhance AI systems by providing more reliable, comprehensive, accurate, and up-to-date knowledge from external sources \cite{schwenk2022okvqa,chen2023can}. 

Heuristic mRAG often relies on predefined retrieval strategies that prioritize grounding across multiple modalities into a single primary modality \cite{gui2022kat,chen2022murag,bonomo2025visual,caffagni2024wiki}, which limits their capacity to deliver precise and contextually rich knowledge. Recent studies \cite{su2024hybrid,wang-etal-2024-rag,li2024benchmarking} propose more adaptable pipelines for knowledge retrieval, utilizing multi-agent cooperation \cite{talebirad2023multi} to harness the model’s intrinsic capacity for knowledge exploration. Despite differences in strategy, all these methods depend on retrieving unstructured external knowledge sources, leading to knowledge redundancy and a lack of precision \cite{peng2024graph}. In contrast, KGs offer a structured format for knowledge storage \cite{ji2021survey}, particularly MMKGs \cite{zhu2022multi}, which can be more efficient support for MLLMs to enhance their response.

\subsection{Multimodal Knowledge Graph}
Compared to KGs, MMKGs incorporate diverse multimodal data, making them more suitable for complex scenarios that require multimodal collaboration. The evolution from conventional KGs to MMKGs has also been extensively explored \cite{chen2024knowledge,lymperaiou2024survey}. Researchers often use densely annotated VQA datasets \cite{krishna2017visual} or web-based resources \cite{bollacker2008freebase,ferrada2017imgpedia} to automatically construct MMKGs \cite{liu2019mmkg,baumgartner2020towards}. Ongoing efforts have demonstrated the effectiveness of them in tasks such as VQA \cite{lin2022retrieval}, image classification \cite{ravi2023vlc}, cross-modal retrieval \cite{zeng2023multi}, \etc. Additionally, some studies focused on embedding MMKGs into feature vectors to enhance the reasoning capabilities of MLLMs, yielding impressive results \cite{lee-etal-2024-multimodal,jiang2024mm}.

Notwithstanding these considerable achievements, current works primarily focus on integrating text and image modalities \cite{lymperaiou2024survey} via web data, which may already embedded in the knowledge of MLLMs. The exploration of additional modalities, such as audio and video, as well as practical applications of recent MLLMs for complex real-world tasks, remains limited. Our meticulously crafted MH-MMKG incorporates multiple modalities alongside complex relations as a knowledge base for a domain-specific task. 

\subsection{Retrieve Knowledge on Graph}
By offering reliable and structured information, retrieval over KGs assists LLMs in preserving better factual accuracy \cite{jin2024large,edge2024local}. Common approaches include retrieving relevant subgraphs \cite{ji2024retrieval,he2024g} or integrating the graph into the model’s learning process \cite{luo2024graph}---techniques that have also been extended to MMKGs to improve MLLMs' task completion capabilities \cite{ishiwatari2020relation,lee-etal-2024-multimodal}. However, training an effective retriever is highly data-intensive, especially for modalities like video. Moreover, adding new KGs into powerful closed-source MLLMs is infeasible. Recent work aims to evoke the model's self-searching ability to autonomously navigate toward the necessary knowledge \cite{sunthink,sunthink,wangdirect}. These approaches typically involve route planning \cite{jiang2023structgpt,sunthink} or self-refinement mechanisms \cite{fang2024karpa,chenplan} to improve retrieval accuracy. We extend these ideas to MMKGs and propose a multi-agent self-searching method for knowledge retrieval.
\section{Datasets}
\label{sec:dataset}
The ``Monster Hunter'' is a popular game series and we choose ``Monster Hunter: World'' as our testbed to explore the potential of MLLMs in tackling the gaps in visual and knowledge using MMKGs.

\subsection{MH-MMKG}
\begin{figure}[t]
\centering
\includegraphics[width=1\columnwidth]{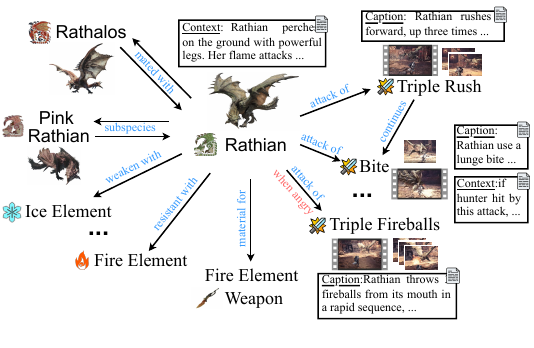}
\vspace{-0.4in}
\caption{The subgraph for monster ``Rathian'' in MH-MMKG.}
\label{fig:kg}
\end{figure}

Our MH-MMKG is an attribute-based MMKG \cite{zhu2022multi} as shown in Figure \ref{fig:kg}, where names of elements that appear in the game, such as monsters and attack actions, are treated as \textit{entities}. A pair of entities can be linked with a particular \textit{relation}, forming an \textit{edge} between them. An entity may come with some additional information, \ie, videos related to the entity and \textit{textual context} to supply more details on the entity, both of which are treated as attributes. Three knowledgeable players were recruited to build the knowledge graph: Each player built subgraphs for single monsters. Figure \ref{fig:kg} illustrates the subgraph for ``Rathian,'' showing intricate relationships such as attack conditions, combos, and inter-subgraph connections. We collected 22 subgraphs, which were merged together to make MH-MMKG.

Let $\mathcal{G} = (\mathcal{E}, \mathcal{V}, \mathcal{R})$ denote our whole KG, where $\mathcal{E}$ is the set of entities, $\mathcal{V}$ is the set of edges, and $\mathcal{R}$ is the set of relations. $\mathcal{E}$ consists of two types of entities: 1) monsters in $\mathcal{E}_\text{o}$ and 2)  other entities in $\mathcal{E}_\text{a}$.
An edge $v = (e, r, e') \in \mathcal{V}$ represents $e \in \mathcal{E}$ has relation $r \in \mathcal{R}$ with $e' \in \mathcal{E}$.  
The attribute associated with an entity $e$ is denoted as $A(e) = (c, u)$, which comprises a video clip $c$ and/or textual context $u$. A video clip describes the entity $e$, recorded in 60 fps and 4k. As current MLLMs are not able to handle videos directly, the dataset also offers human-selected keyframes (no more than 10 frames) as well as human-written captions about $c$.\footnote{The knowledgeable players who curated the dataset also selected the keyframes and wrote the captions.} The textual context $u$ provides knowledge about the entity but is not included in $c$. We denote a set of all attribute as $\mathcal{A} = \{A(e)|e \in \mathcal{E}\}$.

\begin{table}[t]
\caption{The six sub-tasks in our MH benchmark.}
\centering
\resizebox{1\columnwidth}{!}{
\begin{tabular}{lp{7cm}c} 
\toprule
Sub-task & Description & \# samples \\
\midrule
I: Individual Information & Retrieve the textual information of a monster, e.g., nick name, habitat, and skill mechanics. & 24 \\
II: Attack Recognition & Recognize the about to, ongoing or finished attack action of a monster. & 109 \\
III: Combo Premonition & Predict upcoming attack sequences based on the monster’s current action or previous action track. & 28 \\
IV: Condition Awareness & Detect status on monsters or surrounding environment, e.g., whether angry, terrain or phase changes, to anticipate future battle. & 29 \\
V: Proc Effect Insight & Analyze the effects such as environment and attack on monster status and movement patterns. & 35 \\
VI: Cross Monster Analysis & Compare attack patterns or behaviors across different monsters to optimize hunting strategies. & 13 \\
\bottomrule
\end{tabular}}
\label{fig:mission}
\end{table}

\begin{figure*}[t]
\centering
\includegraphics[width=1\textwidth]{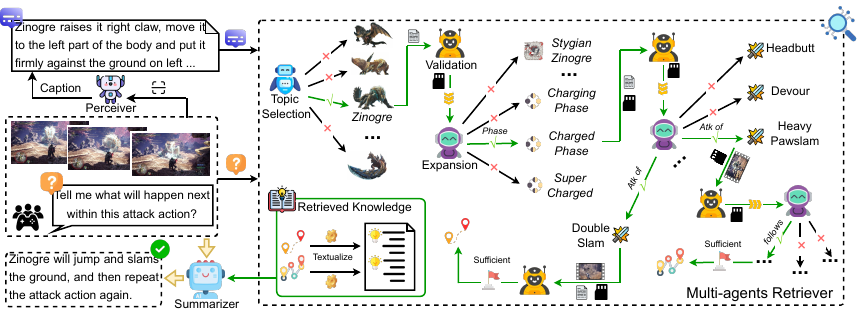}
\caption{Our method first converts the query media into a textual caption. Next, a multi-agent self-search mechanism retrieves relevant knowledge associated with the query. Finally, the retrieved knowledge is utilized to enhance the model's response. \raisebox{-0.2em}{\includegraphics[width=0.3cm]{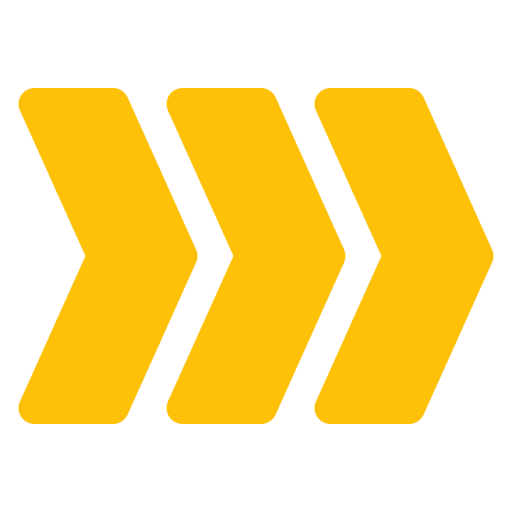}} means the continue of search and \raisebox{-0.2em}{\includegraphics[width=0.3cm]{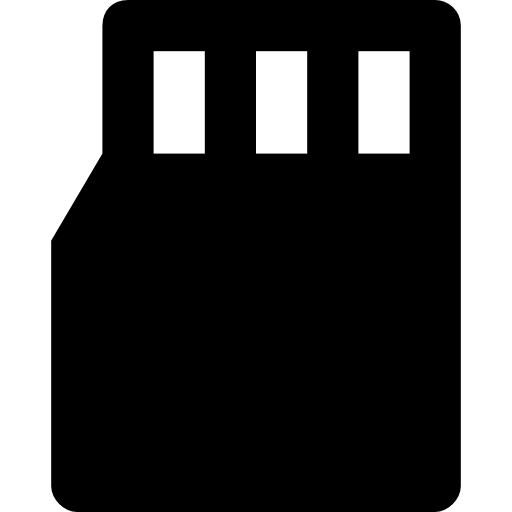}} represents the aggregated knowledge upon current entity.}
\label{fig:method}
\end{figure*}

\subsection{MH Benchmark}

We also designed 238 carefully crafted queries involving knowledge within MH-MMKG. They cover six challenging sub-tasks, as shown in Table \ref{fig:mission} (some samples are available in Figure \ref{fig:sample}). As shown in Figure \ref{fig:method}, each input query $Q = (q, d, z)$ consists of 1) a question $q$, 2) a video or images $d$, which serves as a visual reference for $q$, and auxiliary information $z$, which contains relevant monster's name $e \in \mathcal{E}_o$ as well as additional information that cannot be inferred purely from $d$. The auxiliary information $z$ provides additional context to an MLLM for finding the correct subgraph. By default, we assume that $z$ is available as it facilitates knowledge retrieval and in the real scenarios players generally know them\footnote{We analyze their impacts in the supplementary material.}. Note that, visual data in $Q$ is different from MH-MMKG's. 

To answer $q$ without built-in knowledge about the game's world, a model needs to go through multiple entities and edges of $\mathcal{G}$. For example, to answer question ``Which attack follows?'' with a visual context in which a monster \textit{Rathian} perform attack action \textit{Triple Rush}, the knowledge will be: 
\begin{align}
    \textit{Rathian} \xrightarrow{\text{attack of}}  \textit{Triple Rush} \xrightarrow{\text{continues}} \textit{Bite},
\end{align}
as shown in Figure \ref{fig:kg}. Therefore, finding the correct paths over $\mathcal{G}$ is essential for this task. 
Each query is annotated with a subgraph $\mathcal{I}$ of $\mathcal{G}$ as ground-truth knowledge for answering $q$, textual descriptions $s$ of $d$ written by the knowledgeable game players, and the ground-truth answer $y$. $\mathcal{I}$ comes with a root entity $e_0$, and the paths from $e_0$ to each leaf is deemed as a bunch of knowledge. The details of MH-MMKG and the MH benchmark are available in the supplementary material.

\subsection{Task Definition} \label{task}
We define three task variants over MH benchmark with different levels of supplementary information.

\textbf{Knowledgeable} (Know.) variant simulates a model with sufficient perception for the game's visual domain and sufficient (built-in) knowledge about the world to answer the given question. This idealized setting assesses how well the model reasons the answer $\hat{y}$ from perfect external knowledge and perception. Together with $Q$ and $\mathcal{G}$, we provide a model with 1) the annotated subgraph $\mathcal{I}$ as well as 2) the textual description $s$ of $Q$'s visual reference $d$, and 3) the set of all human-written caption for $c$ in $\mathcal{G}$ so that it can access to ones associated with entities in $\mathcal{I}$ without visual perception, \ie:
\begin{equation}
    \hat{y} = M(Q, s, \mathcal{I}, \mathcal{G}, \mathcal{A}).
\end{equation}

\textbf{Perceptive} variant also assumes a model's sufficient perception but without knowledge to answer the question. This setting evaluates the model's knowledge retrieval (\ie, to find $\mathcal{I}$) ability and the knowledge comprehension ability. The model is supplied with human-written caption for all $c$ so that it can find textual descriptions associated with entities, and is required to output both the final response $\hat{y}$ and the retrieved $\hat{\mathcal{I}}$:
\begin{equation}
    \hat{y}, \hat{\mathcal{I}} = M(Q, s, \mathcal{G}, \mathcal{A}).
\end{equation}

\textbf{Unaided} is the most challenging variant, where the model relies entirely on its own visual perception (for both $d$ and $c$) to interpret the visual world and retrieve external knowledge. As with the perceptive variant, the output are both $\hat{y}$ and $\hat{\mathcal{I}}$. This variant is formulated as:
\begin{equation}
    \hat{y}, \hat{\mathcal{I}} = M(Q, \mathcal{G}).
\end{equation}
\section{Method}
\label{sec:methods}

Figure \ref{fig:method} illustrates the overall pipeline of our method to address our task. It is designed as a baseline method for the MH Benchmark and we describe it for the unaided variant, while it can easily adapted to the others. Given an input query $Q$, the visual reference $d \in Q$ is first converted into a textual caption by a $\textit{perceiver}$. Then, a \textit{multi-agent retriever}, consisting of topic selection, validation, and expansion agents, retrieves relevant knowledge for $Q$. Finally, a \textit{summarizer} generates $\hat{y}$ using retrieved knowledge.

The perceiver $P$ is designed to transform $d$ in $Q$ into textual description. While retaining the original visual data could provide richer information, on-the-fly perception in the visual modality can incur high computational costs; therefore, we choose to convert $d$ into the text as a more efficient representation. The transformed text is demoted as $\hat{s} = P(Q)$. 
We detail the multi-agent retriever and summarizer in the following sections. 

\subsection{Multi-agents Retriever}

We design a fully automated search algorithm with three agents to find the subgraph of $\mathcal{G}$ to answer $q$. First, the topic selection agent $L$ analyzes the input question $q$ to identify the topic entity $e_{0}=L(q, z, \mathcal{E}_o)$, which serves as the root for knowledge retrieval. Then, the expansion and validation agents are alternately activated to grow the subgraph with breadth first search, starting from $e_0$, to include necessary knowledge. The expansion agent finds all plausible neighboring entities for each \textit{open} entities. The validation agent, in turn, validate if the current path from the root to each \textit{open} entities is enough to answer $q$.

We denote the subgraph of $\mathcal{G}$ after the $t$-th round by $\mathcal{K}_t = (\mathcal{E}'_t, \mathcal{V}'_t)$, where $\mathcal{E}'_t \subseteq \mathcal{E}$, $\mathcal{V}'_t \subseteq \mathcal{V}$, and $\mathcal{K}_0 = (\{e_0\}, \emptyset)$. We also denote the set of \textit{open} entities after the $t$-th round by $\mathcal{O}_t \subset \mathcal{E}'_t$, which is a (sub)set of newly added entities at the $t$-th round and is required to explore further expansion.

For the $(t+1)$-th round, the expansion agent $W$ retrieves the set $\mathcal{N}(e)$ of all plausible neighboring entities for each \textit{open} entity $e \in \mathcal{O}_t$ as:
\begin{align}
    \mathcal{N}(e) = W(e, \mathcal{K}_t; Q, \mathcal{G}, \mathcal{A}),
\end{align}
where $W$ is an MLLM with a designated prompt (detailed in the supplementary material) to judge if the knowledge aggregated over the path on $\mathcal{G}$ from $e_0$ to each $e' \in \{e'|(e, e') \in \mathcal{V}\}$ is useful to answer $q$. $\mathcal{N}(e)$ is a set of entities that are judged to be useful. To build $\mathcal{K}_{t+1} = (\mathcal{E}'_{t+1}, \mathcal{V}'_{t+1})$, we first aggregate all neighboring entities and add them to $\mathcal{E}'_{t}$ as:
\begin{align}
    \mathcal{E}'_{t+1} = \mathcal{E}'_{t} \cup \{e'| e \in \mathcal{O}_t, e' \in \mathcal{N}(e)\},
\end{align} 
where duplicated entities are removed to make entities in $\mathcal{E}'_{t+1}$ unique. $\mathcal{V}'_{t+1}$ includes additional links to newly added entities, given by:
\begin{align}
    \mathcal{V}'_{t+1} = \mathcal{V}'_{t} \cup \{ \nu(e, e')| e \in \mathcal{O}_t, e' \in \mathcal{N}(e)\},
\end{align}
where $\nu(e, e')$ gives $v \in \mathcal{V}$ identified by $(e, e')$.

The validation agent $U$ checks if each path from $e_0$ to $e \in \mathcal{O}_{t+1}$ provides sufficient knowledge to answer $q$:
\begin{align}
    o(e) = U(e, \mathcal{K}_{t+1}; Q, \mathcal{G}, \mathcal{A}),
\end{align}
where $U$ again is an MLLM with a carefully designed prompt. $o(e) = \textit{Yes}$ if the knowledge is good enough and no more expansion is needed; otherwise, $o(e) = \textit{No}$ and continues the expansion \footnote{$U$ also output a caption for $c$ of the entity $e$ in unaided-online setting.}.

The open set $\mathcal{O}_{t+1}$ basically is the set of newly added entities in the $(t+1)$-th round. If a newly added entity $e$ is already in $\mathcal{E}'_{t}$ and also in $\mathcal{O}_{t}$, the subgraph forms a loop. In this case, $e$ does not need further expansion as it is already in $\mathcal{E}'_{t}$. Meanwhile, $e$ may be included in the neighbor sets of multiple $e \in \mathcal{O}_t$. In this case, the subgraph also forms a loop, but $e$ has not yet been explored. Thus, it should be in $\mathcal{O}_{t+1}$. Taking these cases into account, $\mathcal{O}_{t+1}$ is given by:
\begin{align}
    \mathcal{O}_{t+1} = \mathcal{E}'_{t+1} \backslash \mathcal{E}'_{t},
\end{align}
where the operator ``$\backslash$'' represents set subtraction. 

This multi-agent retrieval runs until no open entity is found (\ie, $\mathcal{O}_t = \emptyset$). The retrieved subgraph is given by:
\begin{equation}
\hat{\mathcal{I}} = \mathcal{K}_t.
\end{equation}

\subsection{Reasoning via Knowledge Augmentation}
For answer reasoning, we aggregate knowledge on $\hat{\mathcal{I}}$ and represent it in text, which is fed into an LLM. The knowledge augments the reasoning process: the LLM does not need to rely on built-in knowledge about the game's world but (almost) pure reasoning ability is required. This is formally denoted by:
\begin{equation}
\hat{y} = \text{MLLM}(Q, \aleph(\hat{\mathcal{I}}, \mathcal{G}, \mathcal{A}, \alpha)),
\end{equation}
where $\hat{y}$ is the predicted answer for $Q$, and $\aleph$ transforms each path from $e_0$ to each \textit{leaf} of $\hat{I}$ into text given the entire knowledge (\ie, $\mathcal{G}$ and $\mathcal{A}$). The parameter $\alpha$ limited the number of paths in $\hat{\mathcal{I}}$ used for reasoning (defaulted as 5).

\section{Results}
\label{sec:results}

\begin{table}[!t]
	\caption{The experimental settings of query and knowledge retrieval. ``Vision'' indicates that $d$ is used, while ``H-Cap.'' uses $s$. For MMKG, ``Path'' means a model can access annotated subgraphs $\mathcal{I}$. ``H-Cap.'' means a model use human-written captions (for $c$) in $\mathcal{A}$. ``Vis.-Off.'' and ``Vis.-On.'' represent how a model's captions are generated (offline and online).}
	\label{settings}
	\centering
	\resizebox{1\columnwidth}{!}{
    	\begin{tabular}{lcccccc}
    	\toprule
    	  & \multicolumn{2}{c}{Query} & \multicolumn{4}{c}{MMKG} \\
            \cmidrule(lr){2-3} \cmidrule(lr){4-7}
            Methods & Vision & H-Cap. & Path & H-Cap. & Vis.-Off & Vis.-On \\
            \midrule
            Vanilla & \ding{52} &  &  &  & &  \\
            Vanilla$^+$ &  & \ding{52} &  &  & & \\ 
            Know. &  & \ding{52} & \ding{52} & \ding{52} & & \\ 
            Perceptive &  & \ding{52} &  & \ding{52} & & \\ 
            Unaided-Offline & \ding{52} & &  &  & \ding{52} & \\ 
            Unaided-Online & \ding{52} &  &  &  & & \ding{52} \\ 
    	\bottomrule
    	\end{tabular}
	}
    \vspace{-0.1in}
\end{table}

\begin{table*}[t]
\caption{Experimental results are reported for both leading closed-source and open-source MLLMs. The Vanilla, Vanilla$^+$, and Knowledgeable (Know.) experiments are evaluated solely on \textit{Acc.}, while all other experiments are assessed based on both $Acc.$ and knowledge consistency. Results improved in Online than Offline are highlighted with \colorbox{green!20}{light green}.}
\centering
\resizebox{1\textwidth}{!}{
\begin{tabular}{lcccccccccccc}
\toprule
& \multirow{2}{*}{Vanilla} &\multirow{2}{*}{Vanilla$^+$} & \multirow{2}{*}{Know.}& \multicolumn{3}{c}{Perceptive} & \multicolumn{3}{c}{Unaided-Offline} & \multicolumn{3}{c}{Unaided-Online}\\
  \cmidrule(lr){5-7} \cmidrule(lr){8-10} \cmidrule(lr){11-13} 
Models &  &  & & \textit{Acc.} & \textit{Pre.}  & \textit{Rec.} & \textit{Acc.} & \textit{Pre.}  & \textit{Rec.} & \textit{Acc.} & \textit{Pre.}  & \textit{Rec.}\\ 
\midrule
GPT-4o \cite{achiam2023gpt} & .3122 & .3924 & .8565 &\underline{\textbf{.7383}} & .5061 & \underline{\textbf{.7046}} & .4050 & .2595 & .4416 &\cellcolor{green!20}\underline{\textbf{.5105}} &\cellcolor{green!20}.2756 &\cellcolor{green!20}\underline{\textbf{.5625}}\\
GPT-4o mini \cite{achiam2023gpt} & \underline{\textbf{.3218}} & \underline{\textbf{.4135}}  & .8481 & .6877 & .2963 & .5450 & \underline{\textbf{.4514}} & .2059 & \underline{\textbf{.5028}} & .3544 & .1626 & .3009 \\
Claude 3.7 Sonnet \cite{TheC3} &  .2827  & .3375 & \underline{\textbf{.8987}} & .7004 &.5817  & .6322 & .3628 & .2775 & .3270 &\cellcolor{green!20}.4388 &\cellcolor{green!20}\underline{\textbf{.2911}} &\cellcolor{green!20}.4029 \\
Claude 3.5 Sonnet \cite{TheC3} & .2869 & .3755 & .8776 & .7215 &  \underline{\textbf{.5922}} & .6800 & .3966 & \underline{\textbf{.3215}} & .4008 & .3966 & .2330 & .3270 \\
Claude 3.5 Haiku \cite{TheC3} & .2356 & .3206 & .8823  & .6455 & .3739 & .5007 &.3544 &.2002 & .3164 & \cellcolor{green!20}.3670 & .1735 &\cellcolor{green!20}.3361 \\
Gemini 2.0 Flash \cite{team2023gemini}  & .1983 & .2995 & .8438 & .6919 & .3507 & .6146 & .3839 & .1703 & .4092 & .3713 & .1515 & .3663 \\
Gemini 1.5 Pro \cite{team2023gemini} &  .2194  & .2700  & .8438  & .6962 & .4761 & .6033 & .3164 & .1615 & .2194 &\cellcolor{green!20}.4050 &\cellcolor{green!20}.2122 &\cellcolor{green!20}.4585 \\
Step-1o \cite{Stepo1} &  .2436 & .2815 & .8235 &.5747 & .4372 & .5095& .3025 & .1831 & .2483 &\cellcolor{green!20}.3403 &\cellcolor{green!20}.2204 &\cellcolor{green!20}.2987\\
\midrule
InternVL2.5-78B-MPO \cite{chen2024expanding} &  .1603 & .2616 & .8649 &  .5991  & .4198 & .5428 & .2700  & .1729 & .2250 & \cellcolor{green!20}.3080 & .1556 & \cellcolor{green!20}.2378\\
Qwen2.5-VL-72B \cite{bai2025qwen2} & .1476 & .2616 & .8734 & .6244 & .4602 & .4908 &.3206 & .2139 & .2383 &  .3164  & .1615 & .1814 \\
Ovis2-16B \cite{lu2024ovis} & .1645 & .2573 & .8902 & .7046 &  .5407  & .5949 & .2869  & .1963 &.2383 &\cellcolor{green!20}.3459  &.1853 &\cellcolor{green!20}.2878\\
MiniCPM-o-2.6 \cite{yao2024minicpm} & .1139 & .2405 & .8312 & .4683 &  .3183  &  .4001 &.2194 & .1311 & .2376 & .1687 & .0189 & .0210\\
DeepSeek-VL2-Small \cite{wu2024deepseek} & .1139 & .2362 & .6455 & .3586 & .1419 & .1708 & .1814 & .0400 & .0759 & .1181  & .0042 &.0042 \\
\midrule
Human (Knowledgeable) &  .5252  & --- & --- & --- & --- & --- & --- & --- & --- & .9033 & .9207 & .8535 \\
Human (Random) & .0336 & --- & --- & --- & --- & --- & --- & --- & --- & .6092 & .7113 & .6457 \\
\bottomrule
\end{tabular}}
\label{base_line}
\vspace{-0.1in}
\end{table*}

\subsection{Experimental Settings}

We evaluated both leading closed-source models (showing in api: gpt-4o-2024-11-20, gpt-4o-mini-2024-07-18 \cite{achiam2023gpt}, claude-3-7-sonnet-20250219, claude-3-5-sonnet-20240620, claude-3-5-haiku-20241022 \cite{TheC3}, gemini-1.5-pro-002, gemini-2.0-flash-001 \cite{team2023gemini}, and step-1o-vision-32k \cite{Stepo1}) and open-source models (InternVL2.5-78B-MPO \cite{chen2024expanding}, Qwen2.5-VL-72B \cite{bai2025qwen2}, Ovis2-16B \cite{lu2024ovis}, and DeepSeek-VL2 \cite{wu2024deepseek}) with our baseline over MH Benchmark. A single model serves as all agents and functionalities (\ie, the perceiver, summarize, as well as the topic selection, validation, and expansion agents in the multi-agent retriever) via different prompts. We use images solely as visual references because current MLLMs do not support videos as input. Due to input size limitations in most models, all images are resized to 1K for our experiments.

\textbf{Evaluation Metrics.} We use \textbf{accuracy} (\textit{Acc.}) to measure the correctness of predicted answers $\hat{y}$ compared to the ground-truth answer $y$. Since the questions in our benchmark follow an open-ended question-answering format, we employ GPT-4o as a judge with few-shot samples \cite{zheng2023judging}. \textbf{Knowledge consistency} is also used to calculate the consistency between $\mathcal{\hat{I}}$ and $\mathcal{I}$ using precision (\textit{Pre.}) and recall (\textit{Rec.}), demonstrating the efficiency in searching relevant knowledge. Human evaluation for GPT-4o as a judge and metric definition are shown in the supplementary material.

\subsection{Performance of MLLMs}

In addition to the three tasks defined in \ref{task}, we evaluate the models' performance without external knowledge (vanilla). Vanilla$^+$  isolates the effect of the model's visual ability. Furthermore, the Unaided task includes two variants: 1) Offline variant relies on pre-extracted captions (for all $c$ in MMKG) of an MLLM without access to the query $Q$ or any information obtained during knowledge retrieval. Online variant generates captions for $c$ during knowledge retrieval (\ie, in $U$). Table \ref{settings} summarizes the experimental setups. We also include human performance for the vanilla and unaided-online settings. We recruited voluntary participants: One is knowledgeable, and the other has not played the game series. They first answered the questions only with $Q$ and then were allowed to look into $\mathcal{G}$.

As shown in Table \ref{base_line}, with the vanilla setting, all methods hardly predicted correct answers. GPT-4o achieves the best performance with an accuracy of 0.31, whereas the knowledgeable human attains 0.53. The closed-source models generally outperform open-source ones, implying that the closed-source models have richer built-in knowledge about the game. With the vanilla$^+$ setting, which allows human captions $s$ instead of images $d$, most models' performance improved, which means that MH Benchmark requires comprehension of visual context. In the knowledgeable experiment, the models achieved an accuracy of roughly 0.9. We thus argue that MH-MMKG provides sufficient knowledge to answer the questions in MH Benchmark.

\begin{figure}[t]
\centering
\includegraphics[width=0.9\columnwidth]{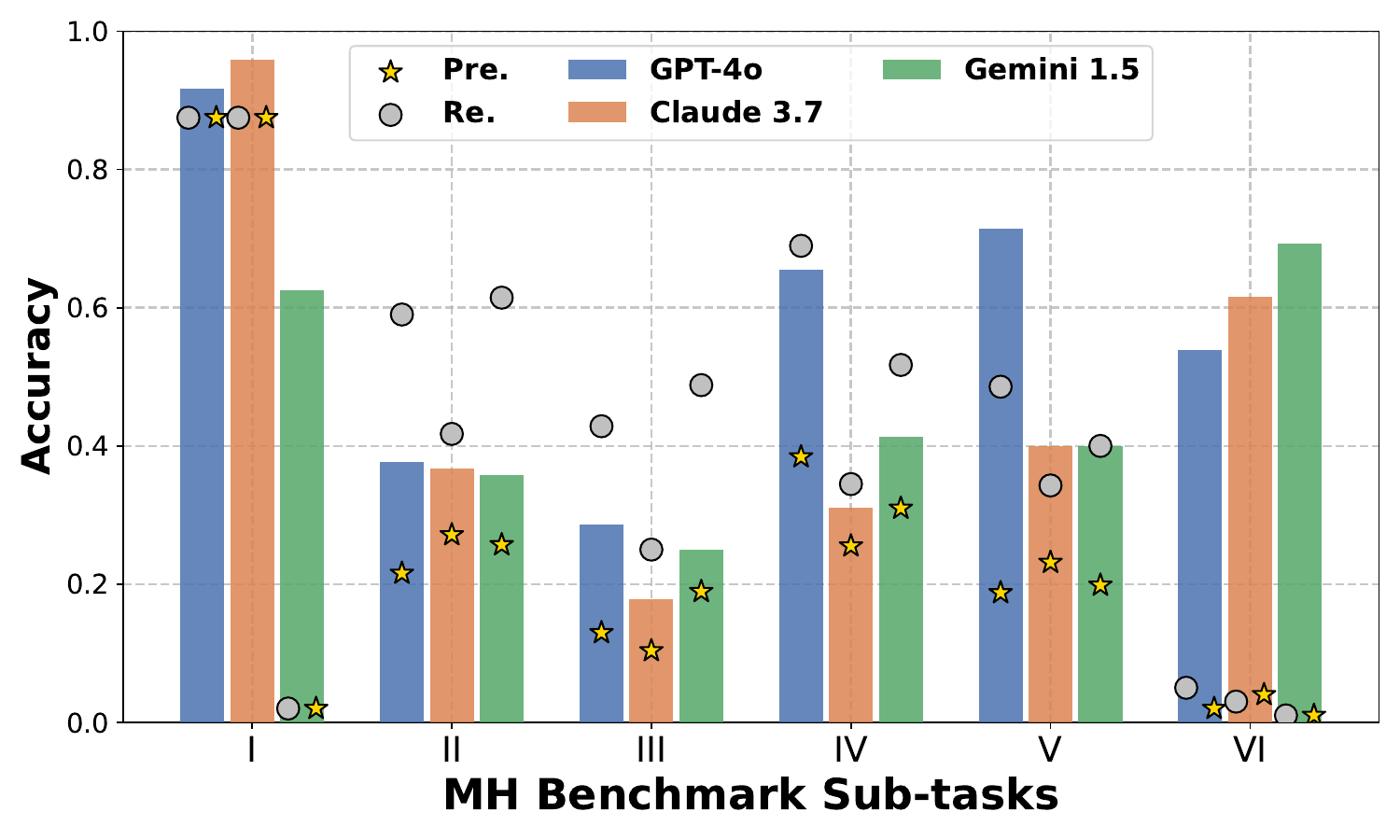}
\caption{Performance comparison of GPT-4o, Claude 3.7 Sonnet, and Gemini 1.5 Pro across 6 sub-tasks in MH Benchmark.}
\label{fig:mission2}
\vspace{-0.1in}
\end{figure}

\begin{figure*}[t]
\centering
\includegraphics[width=0.95\textwidth]{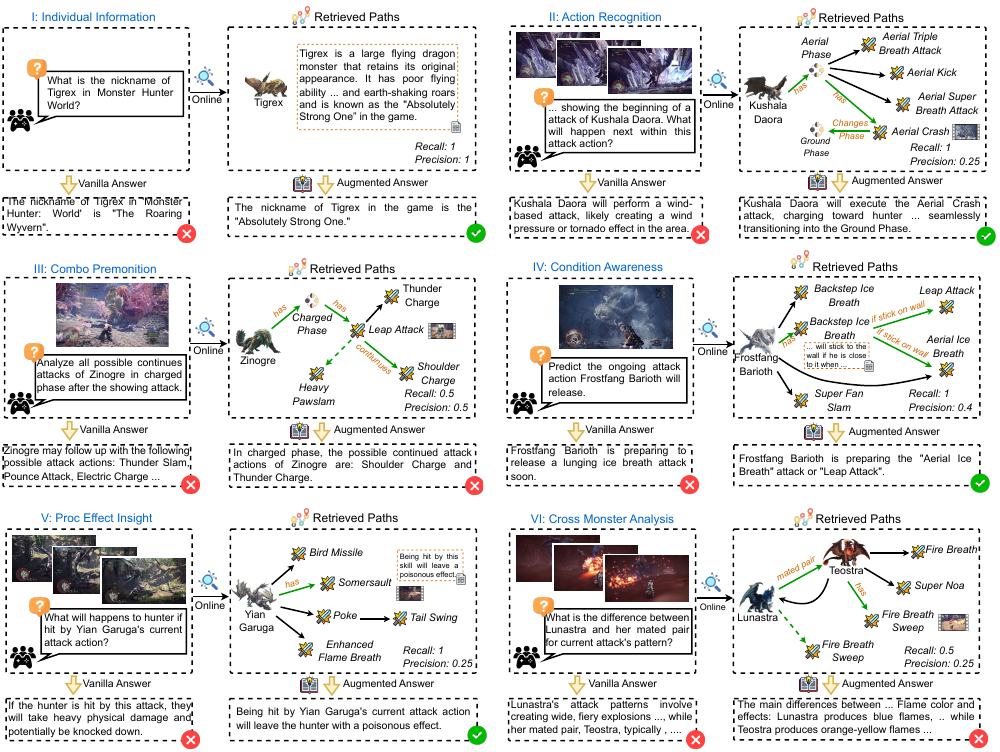}
\caption{Examples of the 6 sub-tasks in the MH Benchmark, each generated by GPT-4o for both the Vanilla Answer and the Augmented Answer using an unaided-online retrieval.}
\label{fig:sample}
\end{figure*}

The perceptive, unaided-offline, and unaided-online settings evaluate a model's ability to retrieve relevant knowledge In the perceptive setting, which uses human captions for both queries and knowledge, all models demonstrated a strong capacity to identify relevant knowledge, significantly improving performance compared to vanilla$^+$. Additionally, the models exhibit high knowledge consistency. In the unaided-offline setting, all models performed worse than in the perceptive setting, suggesting that visual perception plays a crucial role in knowledge retrieval. Nevertheless, they are still still better than the vanilla setting, demonstrating the benefits of knowledge retrieval. The unaided-online setting further challenges the models' visual perception to generate captions with richer context (\ie, $Q$ and the knowledge on the path in $\hat{\mathcal{I}}$) compared to the unaided-offline. Our results show that only some models surpassed the offline variant (highlighted in green), yet the improvement is evident--GPT-4o improves by 0.1, and Claude 3.7 Sonnet by 0.07. This suggests that \textit{knowing what to do} enhances a model's visual perception, strengthening its planning ability, and ultimately results in improved reasoning ability. Additionally, we find that recall is consistently higher than precision, suggesting that models tend to retrieve more knowledge. On the contrary, humans showed higher precision (though both precision and recall are high compared to MLLMs).

We also analyzed the performance differences in the unaided-online setting across six sub-tasks over GPT-4o, Claude 3.7 Sonnet, and Gemini 1.5 Pro, as illustrated in Figure \ref{fig:mission2}. GPT-4o achieves the best performance in all sub-tasks except I and VI. It is particularly strong in sub-tasks IV and V, which are more reliant on fine-grained visual perception. Additionally, sub-task I is the simplest, involving only a single relationship $v$, for which the models generally perform well.  For sub-task VI, although the accuracy is high, models fail to find the correct path (model uses its inherent knowledge for response). These results suggest that visual perception remains a key challenge for MH benchmark.

Figure \ref{fig:sample} presents six examples (one per sub-task) in the vanilla and unaided-online settings, where all predictions are by GPT-4o. In the retrieved path panes, the green paths are both in $\mathcal{I}$ and $\hat{\mathcal{I}}$ (\ie, correctly retrieved paths), the dotted green paths are in $\mathcal{I}$ but not in $\hat{\mathcal{I}}$, and the black paths are in $\hat{\mathcal{I}}$ but not in $\mathcal{I}$. The key relationship or essential information for deriving the correct answer is highlighted in orange. We report the precision and recall metrics for each example. GPT-4o retrieves many knowledge paths, capturing all relevant knowledge for prediction, though precision is low. Interestingly, for challenging cases like sub-tasks II, GPT-4o successfully recognized detailed visual concepts (\eg, the monster flying backward and the monster sticking to the wall) and generated plausible captions, which require fine-grained visual perception.

\subsection{Analysis of Factors Affecting Performance}

\textbf{Impact of Captioning Ability.} Table \ref{caption_experiments} compares performance of captioning, where the metric \textit{similarity} (Sim.) measures the similarity between generated and human captions using GPT-4o (see supplementary for details). The table also summarizes the accuracy and knowledge consistency scores. The results indicate that the unaided-online setting consistently gives higher similarity scores than the unaided-offline, suggesting that awareness of query $Q$ and retrieval process consistently enhances captioning performance. Also, the similarity scores positively correlated with the reasoning accuracy scores. We also evaluated the unaided-offline variant on GPT-4o but its offline captions are replaced with ones by video models InternVideo2.5 \cite{wang2025internvideo2} and VideoChat-Flash \cite{li2024videochat}. The results showed lower performance than the original GPT-4o across all metrics, indicating that current video models have limited capability in visual comprehension for MH Benchmark.


\begin{table}[!t]
    \caption{Reasoning and knowledge consistency performance and captioning performance. Improved performance scores due to online captioning are highlighted in green.}
    \centering
    \resizebox{0.98\columnwidth}{!}{
    \begin{tabular}{lcccccc}
        \toprule
        \textbf{Model} & \textbf{Vis.-Off} & \textbf{Vis.-On} & \textit{Acc.} & \textit{Pre.} & \textit{Rec.} & \textit{Sim.} \\
        \midrule
        GPT-4o \cite{achiam2023gpt} & \ding{52} & & .4050 & .2595 & .4416 & .2806\\
        GPT-4o \cite{achiam2023gpt}  &  & \ding{52} & .5105 & .2756 & .5625 & \cellcolor{green!20}.2948\\
        Claude 3.7 Sonnet \cite{TheC3} & \ding{52} & & .3628  & .2775 & .3270  & .2776 \\
        Claude 3.7 Sonnet \cite{TheC3} &  & \ding{52} & .4388 & .2911 & .4029 &\cellcolor{green!20}.3208\\
        Gemini 1.5 Pro \cite{team2023gemini} & \ding{52} & & .3164  & .1615  & .2194  &.1608\\
        Gemini 1.5 Pro \cite{team2023gemini} &  & \ding{52} & .4050 & .2122 & .4585 &\cellcolor{green!20}.1746\\
        \midrule
        InternVideo2.5 \cite{wang2025internvideo2} & \ding{52} & & .3697 & .1960 & .2959 &  .0525\\
        VideoChat-Flash \cite{li2024videochat} & \ding{52} & & .3445 & .2135 & .2863 & .0644 \\
        \bottomrule
    \end{tabular}
    }
    \label{caption_experiments}
\vspace{-0.1in}
\end{table}

\textbf{Keyframe Selection.} Due to the limited number of input tokens, current MLLMs cannot handle videos, and keyframe sampling is necessary. MH-MMKG provides human-selected keyframes to facilitate the evaluation on this point. To show the difference between human-selected keyframes and typically-adopted equal-interval sampling, Table \ref{key_frame} summarizes the results on GPT-4o, where sampling is done in two frames per second, and the maximum number of frames is capped at ten frames. The keyframe selection strategy does impact the performance: Human-selected keyframes seem to provide more informative visual cues, though the difference is not substantial. 

\begin{table}[!t]
    \caption{Impact of Keyframes selection.}
    \centering
    \resizebox{0.80\columnwidth}{!}{
    \begin{tabular}{lcccccc}
        \toprule
         &  \multicolumn{3}{c}{Unaided-Offline} & \multicolumn{3}{c}{Unaided-Online}\\
        \cmidrule(lr){2-4} \cmidrule(lr){5-7}
        Settings & \textit{Acc.} & \textit{Pre.}  & \textit{Rec.} & \textit{Acc.} & \textit{Pre.}  & \textit{Rec.}\\ 
        \midrule
        Human  & .4050 & .2595  & .4416 & .5105 & .2756 & .5625\\
        Sampling & .3725 & .2199 & .3893 & .4840 &.2388 & .5254 \\
        \bottomrule
    \end{tabular}
    }
    \label{key_frame}
    \vspace{-0.1in}
\end{table}

\begin{figure}[t]
    \centering
    \begin{subfigure}{0.575\linewidth}
        \centering
        \includegraphics[width=\linewidth]{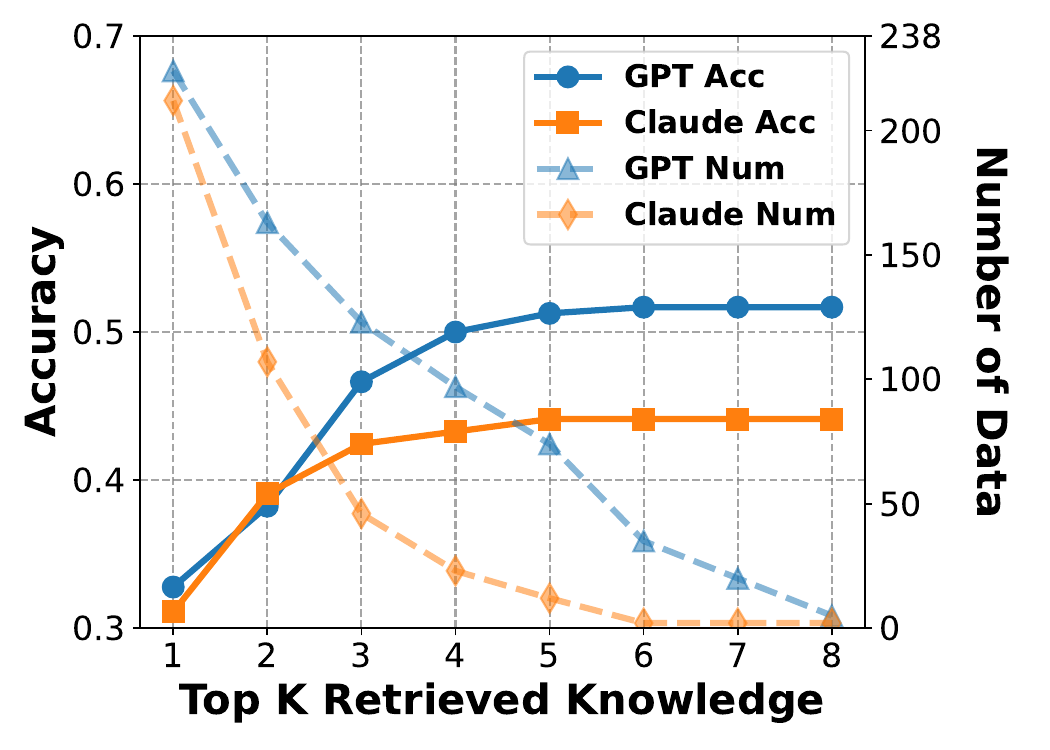}
        \caption{Using K paths as knowledge.}
        \label{fig:a}
    \end{subfigure}
    \hfill
    \begin{subfigure}{0.415\linewidth}
        \centering
        \includegraphics[width=\linewidth]{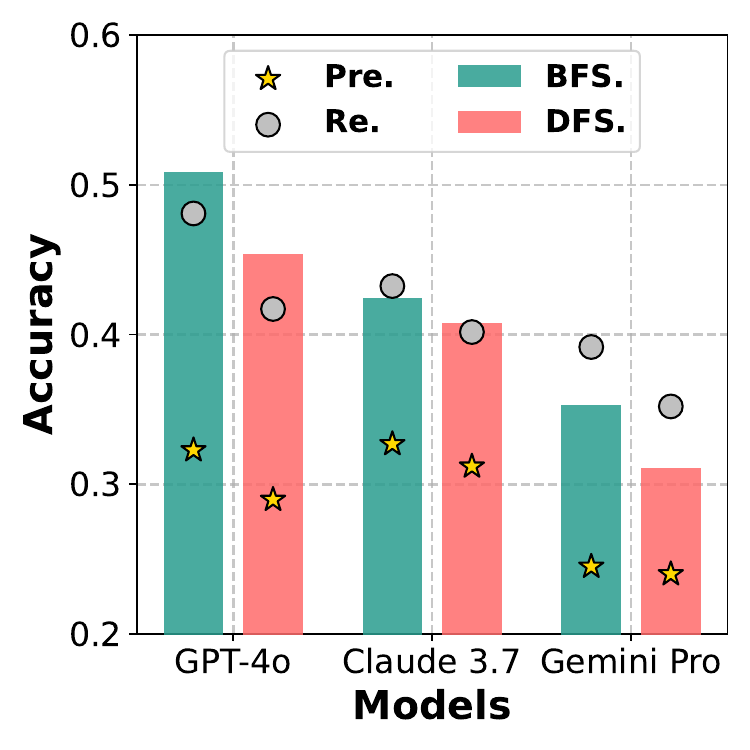}
        \caption{BFS vs DFS}
        \label{fig:b}
    \end{subfigure}
    \caption{Ablation experiments for proposed multi-agents search.}
    \label{fig:subfigures}
    \vspace{-0.1in}
\end{figure}

\textbf{The Number of Paths Used in Reasoning.} In our experiments, the number of the paths in $\hat{\mathcal{I}}$ used for reasoning is limited. All evaluations so far used 5 paths, though this number can change the performance \footnote{Our baseline uses the first 5 paths. A shorter path is thus preferred.}. Figure \ref{fig:a} shows the relationship between the number of paths and the reasoning accuracy. GPT-4o gives the optimal performance when 5 paths are used. We also show in the figure the number of queries $Q$ for which at least a certain number of paths are retrieved. GPT-4o tends to find more paths as its knowledge (the recall is higher), while Claude 3.7 is more conservative in retrieval (the precision is higher).

\textbf{BFS versus DFS.} The retrieval strategy can also impact performance. BFS finds shorter paths first, while depth-first search (DFS) can find longer paths at the early stage of retrieval. We evaluated the performance for BFS (the proposed baseline) and DFS when three paths are used. As shown in Figure \ref{fig:b}, BFS consistently performs better. This means that our queries can be answered within fewer hops in $\mathcal{G}$, which can be seen as a limitation of our MH Benchmark, as it only requires fewer steps of reasoning.
\section{Conclusion}
\label{sec:conclusion}
This work explores the ability of MLLMs to handle domain-specific tasks by retrieving knowledge from MMKG. We introduce MH-MMKG and the MH benchmark as a testbed. Additionally, we propose a baseline with a multi-agent knowledge retriever that allows MLLMs to autonomously access relevant knowledge without requiring additional training. Experimental results on both highlight the importance of visual perception of MLLMs and finding relevant knowledge for reasoning. Our work paves the way for more adaptable and context-aware MLLMs in complex real-world scenarios. Future research may focus on expanding the benchmark size, incorporating a wider range of knowledge types, exploring the potential of video modalities, and developing more advanced knowledge retrieval methods.


\paragraph{Acknowledgement} This work is supported by World Premier International Research Center Initiative (WPI), MEXT, Japan. This work is also supported by JST ACT-X Grant No.~JPMJAX24C8, JSPS KAKENHI No.~24K20795, CREST Grant No.~JPMJCR20D3, and JST FOREST Grant No.~JPMJFR216O. 

{
    \small
    \bibliographystyle{ieeenat_fullname}
    \bibliography{main}
}

\clearpage
\setcounter{page}{1}
\maketitlesupplementary
\tableofcontents

\section{Detail of MH Benchmark Construction}
In this section, we detailed construction of our MH-MMKG and MH benchmark.

\subsection{MH-MMKG}
A total of 22 monsters are incorporated into the graph construction, with each represented as a subgraph connected through various relationships, such as species relation and elemental weaknesses. Each subgraph contains rich information crucial for successful conquests, particularly regarding attack strategies, combos, attack phases, and launch conditions. The monsters are: Anjanath, Azure Rathalos, Barroth, Bazelgeuse, Brachydios, Diablos, Frostfang Barioth, Glavenus, Kushala Daora, Legiana, Nergigante, Rathalos, Rathian, Teostra, Tigrex, Uragaan, Zinogre, Pink Rathian, Yian Garuga, Stygian
Zinogre, and Radobaan from Monster Hunter World.

\begin{figure}[t]
    \centering
    \includegraphics[width=1\columnwidth]{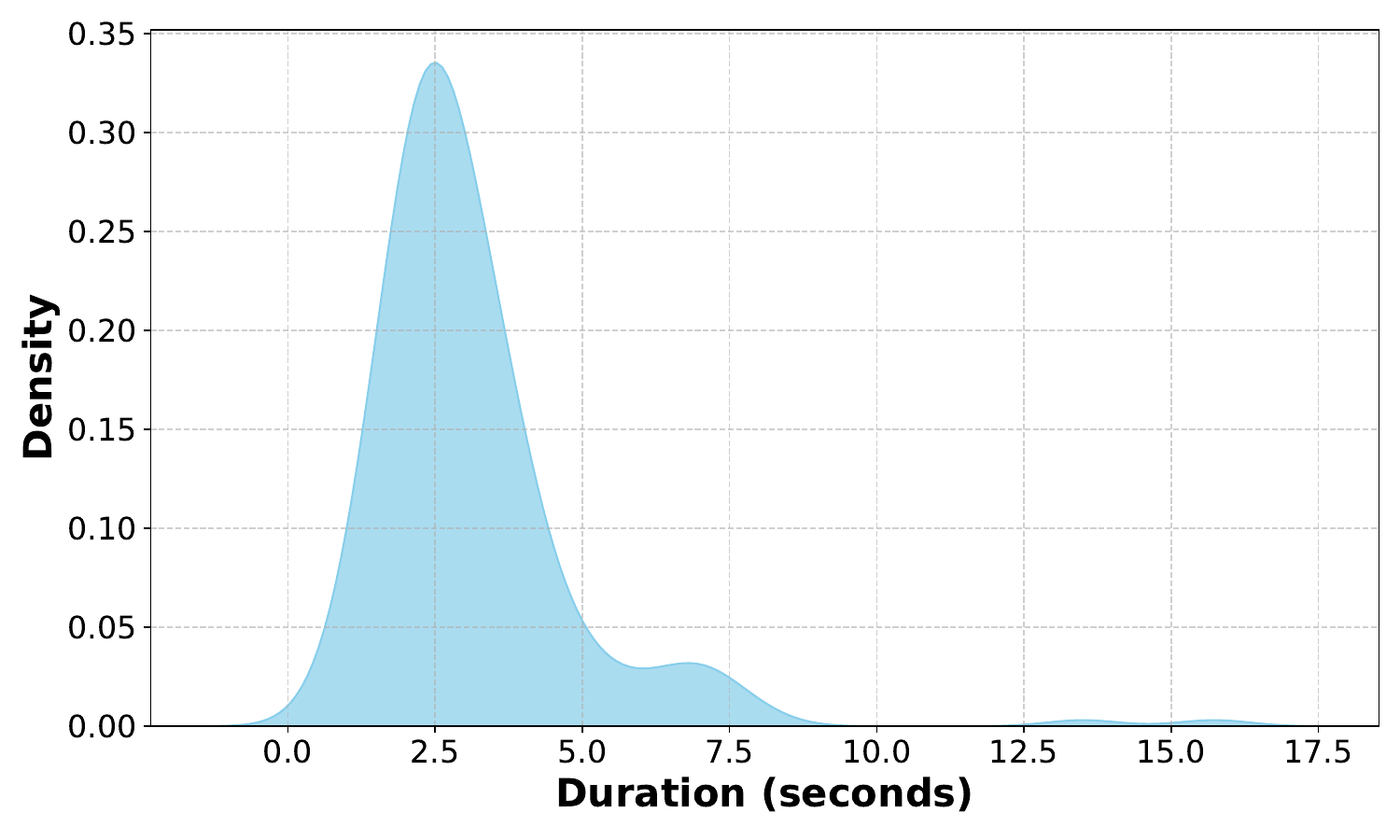}
    \caption{Video clip length statistic.}
    \label{fig:clip}
\end{figure}

To ensure the quality, we hired three experienced Monster Hunter World players, each with over 200 hours of game play experience. They were tasked with gathering relevant monster information from sources such as Wiki, YouTube, and Bilibili to construct the graph. Additionally, since each monster has unique characteristics within the game, the structure of each subgraph is tailored accordingly. The entities are classified into 7 types as show in Table \ref{fig:entity}. Most of them are attack actions, making MH-MMKG more focused on battles with monsters. We also plan to explore more game elements in the future. Some entities are attached with text, image or video as its attribution. Note that all video or images are captured from Arena field. While for queries in MH Benchmark all visual media are captured from the Wild field. We also show the length statistic of video clips in Figure \ref{fig:clip}. It can be observed that most videos are around 1s to 5s.

\begin{table}[t]
\caption{Types of entity in MH-MMKG.}
\centering
\resizebox{1\columnwidth}{!}{
\begin{tabular}{lp{7cm}c} 
\toprule
Type & Description & \# number \\
\midrule
Topic Entity & Names of monsters that can serve as root entities for knowledge retrieval. Each entity is accompanied by an image of monster as its attribute. & 22 \\
Attack Action & Possible attack movements of a monster, each accompanied by text, images (key frames for video), or a video as its attribute. Each of them also attached with human written-caption for the video. & 265\\
Attack Phase & In different phases, a monster will have varying attack patterns, damage, combos, and other attributes. Only some monsters have unique phase settings. Textual context is attached as attribution. &  20  \\
Element & The element indicates a monster's weakened resistance to a specific type of attack. & 9 \\
Weapon & Types of damage for weapons crafted from monster materials. & 10 \\
Props & Various types of game props for interacting with monsters. & 6 \\
Attack Effects & The effects of monster attacks or skills during battle, including generated ice patches on ground, scratches, and explosions.  Textual context is attached as attribution. & 9 \\
\bottomrule
\end{tabular}}
\label{fig:entity}
\end{table}

There are also 158 kinds of edges and  16 of them are base edges: ``has attack action of", ``continues with attack action of", ``has attack variant of", ``has attack phase of", ``change attack phase to", ``is mostly weaken with", ``is weaken with", ``is resistant with",  ``provide materials for",  ``can be stopped by",  ``has attack variants of", ``generates", ``cause", ``turns to", ``mated pair with", ``has subspecies of". Some base edges (mostly the first two of them) are further combined with specific constrain mechanism to form 142 variants. The samples of constrains are: ``is angry", ``hunter step into",  ``is close to", ``stick on the wall", ``is knocked by", etc. (We do not show all of them as they too many ).

\begin{figure}[t]
    \centering
    \includegraphics[width=1\columnwidth]{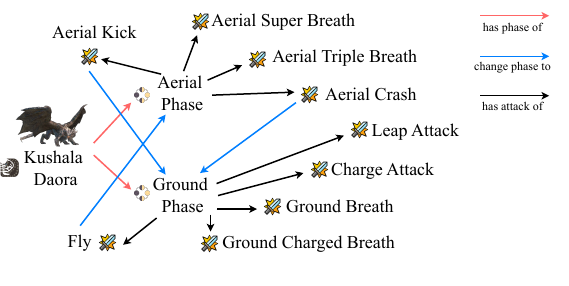}
    \caption{Subgraph structure for Kushala Daora.}
    \label{fig:sub1}
\end{figure}

\begin{figure}[t]
    \centering
    \includegraphics[width=1\columnwidth]{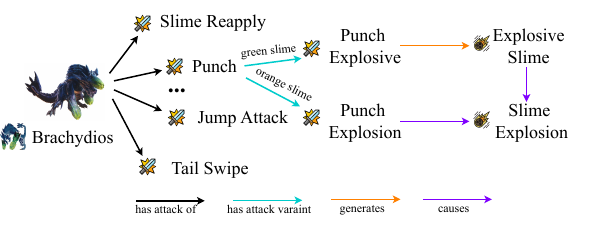}
    \caption{Subgraph structure for Brachydios.}
    \label{fig:sub2}
\end{figure}

We present some sub-graphs to illustrate structural diversity, focusing only on attack actions and their related entities, as the other components resemble Figure 2 in the main paper. The main paper showcases the graph structure of Zinogre, known for its extensive combo attacks. Here, we provide two additional examples: Kushala Daora and Brachydios. Kushala Daora exhibits distinct attack patterns in its Aerial Phase (attacking from the air) and Ground Phase (attacking on the ground), as shown in Figure \ref{fig:sub1}. Certain attacks can transition between these phases, making this information crucial for an MLLM to accurately answer related questions. Brachydios, on the other hand, has attack variations that depend on the color of the slime on its fists or head, as illustrated in Figure \ref{fig:sub2}. The color change alters both the attack variant and its effect, adding another layer of complexity to its combat behavior. MLLMs have to comprehend such complex information to correctly answer the question in MH Benchmark.

\subsection{MH Benchmark}
To differentiate from MH-MMKG, all visual media for queries are captured from the Wild field. Additionally, we present statistics on the average number of entities and depth of knowledge associated with each query in the MH Benchmark, as shown in Table \ref{fig:bench}. It can be observed that sub-task I is relatively simple, as it relies solely on the topic node. In contrast, sub-tasks II and VI involve a greater number of steps and deeper analysis, as they pertain to combo recognition and cross-monster comparison, both of which require more complex reasoning.

\begin{table}[t]
\caption{Average number of entities and depth of knowledge for each query in MH Benchmark.}
\centering
\resizebox{0.9\columnwidth}{!}{
\begin{tabular}{lcccccc} 
\toprule
Sub-tasks & I & II & III & IV & V & VI \\
\midrule
Number$_{avg}$ & 1 & 2.339 & 3.535 & 2.4137 & 3.028 & 4.076 \\ %
Depth$_{avg}$ & 1 & 2.278 & 3.250 & 2.4137 & 2.900 & 3.038 \\
\bottomrule
\end{tabular}}
\label{fig:bench}
\end{table}

\begin{table}[t]
\centering
\caption{Prompt for \textit{perceiver} agent.}
\resizebox{\columnwidth}{!}{
\begin{tabular}{p{10cm}}
\toprule
\textbf{Input Prompt} \\
\midrule
You are a professional Monster Hunter player. You are playing `Monster Hunter: World'.\\
You will receive consecutive video frames displaying the battle screen with the monster \{\textcolor{blue}{monster name}\}. \\
The given `Question' regarding the battle screen is: \{\textcolor{red}{question}\}  \\
Generate a `Description' of the battle scene as your `Response', detailing the monster’s limb and body movements, mouth actions, surroundings, and other relevant details.  \\
Note that you should not give any assumptions for the `Description'.\\
Note that you should directly output your `Response' and do not output any information other than your `Response'.\\
Now, start to complete your task.\\
Your `Response': \\
\bottomrule
\end{tabular}
}
\label{perception}
\vspace*{-4mm}
\end{table}

\begin{table}[t]
\centering
\caption{Prompt for \textit{topic entity selection} agent.}
\resizebox{\columnwidth}{!}{
\begin{tabular}{p{10cm}}
\toprule
\textbf{Input Prompt} \\
\midrule
You are a professional Monster Hunter player. You are playing `Monster Hunter: World'.\\
You will receive consecutive video frames displaying the battle screen with the monster: \{\textcolor{blue}{monster name}\}. \\
The given `Question' regarding the battle screen is: \{\textcolor{red}{question}\}  \\
All possible monster names `Options' are structured in a list format as follows:\{\textcolor{red}{topic entity}\}\\
Note that your `Response' is to directly output the name of the monster you are looking for. \\
Note that you should not output any information other than your `Response'.\\
Now, start to complete your task.
Your `Response':\\
\bottomrule
\end{tabular}
}
\label{selection}
\vspace*{-4mm}
\end{table}

\begin{table}[t]
\centering
\caption{Prompt for \textit{expansion} agent.}
\resizebox{\columnwidth}{!}{
\begin{tabular}{p{10cm}}
\toprule
\textbf{Input Prompt} \\
\midrule
You are a professional Monster Hunter player. You are playing `Monster Hunter: World'.\\
The text description of the battle screen is: \{\textcolor{red}{caption}\}. \\
Based on the battle screen, here is the `Question' you need to answer: \{\textcolor{red}{question}\}. \\
To answer the above question, you are now searching a knowledge graph to find the route towards relevant knowledge. The following contents are the knowledge you found so far (up to current entity \{\textcolor{red}{entity}\}):\\
******\\
\{\textcolor{red}{memory}\}\\
******\\
You need to select the relevant `Neighbor Entity' that may provide knowledge to answer the question. The relation and condition from current entity '{entity}' to all `Neighbor Entity' are:\\
******\\
\{\textcolor{red}{neighbor entity}\}\\
******\\
Your `Response' is directly output the name of all relevant `Neighbor Entity' and separate them directly by `;'.\\
If there is no relevant `Neighbor Entity', directly output 'None'.\\
Note that if the `Neighbor Entity' is an attack action, always choose it (if it is not highly irrelevant).\\
Note that if the `Neighbor Entity' is a phase, you can only choose one.\\
Note that you should not output any information other than your `Response'.\\
Now, start to complete your task.\\
Your `Response':\\
\bottomrule
\end{tabular}
}
\label{expansion}
\vspace*{-4mm}
\end{table}

\section{Experiment Details}
In this section, we show the detailed settings of our baseline method, including prompt for each agent, additional experiments, and more samples.

\subsection{Prompt Template for Our Method}
We first present the prompt templates for all agents in the retrieval pipeline. Table \ref{perception} shows the prompt for the \textbf{perceiver} agent, which translates input images into text based on the given question. Table \ref{selection} provides the prompt for the \textbf{topic selection} agent, responsible for selecting the starting entity for knowledge retrieval from the graph. Table \ref{expansion} contains the prompt for the \textbf{expansion} agent, which plans the next neighboring entity for search. Table \ref{validation} presents the prompt for the \textbf{validation} agent, designed to assess the efficiency of knowledge transfer from the starting entity to the current entity. Finally, Table \ref{summary} includes the prompt for the \textbf{summarizer} agent, which synthesizes the retrieved knowledge for final answer generation. Among these, the \{monster name\}, displayed in blue text, represents additional information as a part of question. The \{entity\} represents the name of current entity during search. The \{question\} refers to the input query, while \{topic entities\} denote the names of all topic entities. \{entity infp\} is the visual irrelavent additional information for an entity. The \{caption\} is the generated description by the \textbf{perceiver} agent. 

The \{neighbor entity\} are options of neighbor for current entity. It is presented in a text format consisting of a combination of entity-edge triplets and corresponding constraints or conditions (if any). Here is a neighbor sample for monster ``Frostfang Barioth" attack action entity ``Straight Ice Breath":
\begin{compactitem}
\item``Straight Ice Breath" continues with attack action of ``Super Fang Slam" (Condition: When hunter hitted by the breath...)
\item``Straight Ice Breath" continues with attack action of ``Tail Spin" (Condition: When Frostfang Barioth already released two...)
\end{compactitem}
In our prompt, we instruct the model to select relevant neighboring entities while placing greater emphasis on attack action entities, as most tasks are designed around them. For phase entities, we allow the model to select only one, in accordance with the game mechanics.

The \{memory\} records the search path from the starting entity to the current entity, including entity names and all relevant information at each step. Below is an example illustrating this transition from a knowledge path: \\
\begin{align}
    \textit{Zinogre} \xrightarrow{\text{phase of}}  \textit{Charged Phase} \xrightarrow{\text{attack of}} \textit{Double Slam}
\end{align}
will be transferred into: 
\begin{compactitem}
\item``Zinogre": Additional Information: Zinogre has the appearance of a wild wolf and lives in the mountains full of dense trees ... 
\item``Zinogre" has attack phase of "Charged Phase". 
\item``Charged Phase": Additional Information: Zinogre is charged, the body will be surrounded by electric ... 
\item``Charged Phase" has attack action of ``Double Slam". 
\item``Double Slam": Action Description: Zinogre lowers his head and rubs the ground with... 
\end{compactitem}

\begin{table}[t]
\centering
\caption{Prompt for \textit{validation} agent. Content in \textcolor{green}{[]} is used solely for unaided-online experiments.}
\resizebox{\columnwidth}{!}{
\begin{tabular}{p{10cm}}
\toprule
\textbf{Input Prompt} \\
\midrule
You are a professional Monster Hunter player. You are playing `Monster Hunter: World'.\\
The text description of the battle screen is: \{\textcolor{red}{caption}\}. \\
Based on the battle screen, here is the `Question' you need to answer: \{\textcolor{red}{question}\}. \\
To answer the above question, you are now searching a knowledge graph to find the route towards relevant knowledge. \\
You are a professional Monster Hunter player. You are playing `Monster Hunter: World'.\\
To answer the above question, you are now searching a knowledge graph to find the route towards relevant knowledge. The following contents are the knowledge you found so far (up to current entity \{\textcolor{red}{entity}\}):\\
******\\
\{\textcolor{red}{memory}\}\\
******\\
And here is some information of current entity: \{\textcolor{red}{entity info}\}. \\
\textcolor{green}{[}You will also receive consecutive video frames showing the battle screen with the monster \{\textcolor{blue}{monster name}\} as visual information for current entity \{\textcolor{red}{entity}\}. \\
Make a `Description' (do not affected by previous text description of the battle screen for the `Question') for the battle screen as a part of your `Response'. `Description' should include monster's limb and body movements, mouth, surrounding and others details. \\
Note that you should not give any assumptions for the `Description'.\textcolor{green}{]}\\
You have to decide whether visual and text information of this entity together with previous found knowledge is sufficient for answering this `Question'. \\
For sufficient analysis, your `Answer' is `Yes' or `No'. \\
\textcolor{green}{[}Directly output your `Response' as the combination of `Answer' and `Description', separating them directly by `;'. \textcolor{green}{]}\\
Note that you should not output any information other than your `Response'. \\
Now, start to complete your task.\\
Your `Response':\\
\bottomrule
\end{tabular}
}
\label{validation}
\vspace*{-4mm}
\end{table}

\begin{table}[t]
\centering
\caption{Prompt for \textit{summarizer} agent.}
\resizebox{\columnwidth}{!}{
\begin{tabular}{p{10cm}}
\toprule
\textbf{Input Prompt} \\
\midrule
You are a professional Monster Hunter player. You are playing `Monster Hunter: World'.\\
You will receive consecutive video frames displaying the battle screen with the monster \{\textcolor{blue}{monster name}\}.
Based on the battle screen, here is the `Question' you need to answer:  \{\textcolor{red}{question}\}. \\
Here is the `Knowledge' you retrieved from a knowledge graph for this `Question':\\
******\\
\{\textcolor{red}{knowledge}\}\\
******\\
Your `Response' is to provide the answer for this `Question' based on the retrieved Knowledge.\\
Note that you should not give any analysis. \\
Note that you should not output any information other than your `Response'. \\
Now, start to complete your task.\\
Your `Response': \\
\bottomrule
\end{tabular}
}
\label{summary}
\vspace*{-4mm}
\end{table}

Note that Additional Information is the attribution of an entity (if exist). Action Description is given as human-made caption in Knowledgeable experiments, pre-extracted from visual attribution (if exist) in unaided-offline, and dynamic generated for visual attribution in unaided-online (if exist). Especially, the \textcolor{green}{[]} highlights the content for unaided-online that requires the model to comprehend visual references during validation and output corresponding description as the temporal visual attribution for the current entity.

As shown in Table \ref{summary}, the final agent \textbf{summary} will treat all retrieved paths as \{knowledge\} using the same strategy as \{memory\}. Each path will be converted into a text description and attached to the query as input.

Table \ref{off} shown the prompt template for unaided-offline experiments. It is used to pre-extract the visual reference (images or video for MLLMs or Video models, respectively in our experiments) into text description. This transition is not related to query or search memory. 

Note that, the prompts for the agent pipeline were developed using InternVL2.5-78B \cite{chen2024expanding}, with the expectation that even open-source models, by their instruction-following capabilities, can understand these prompts and generate responses in the required format. This ensures a fair comparison for all close-source models in the main paper. We further conducted a preliminary prompt robustness analyses for GPT-4o and Claude 3.7 (unaided-online). Our observations show that Claude generally exhibited robust performance across prompt variations, particularly for agents with straightforward instructions such as \textit{Perceiver}, \textit{Topic Selection}, and \textit{Summarizer}. However, GPT-4o exhibited sensitivity to lexical choice. For instance, in the \textit{Validation} agent, the use of the term ``sufficient” to determine whether the retrieved knowledge is enough and the retrieval should be stopped. When we replaced it with ``necessary,” GPT-4o tended to more cautious during retrieval. This minor change led to a .0546 and .0871 drops on \textit{Acc.} and \textit{Rec.}, respectively, though with a .0194 improvement in \textit{Pre.} These findings suggest that prompt robustness is both model-specific and agent-specific.

\begin{table}[t]
\centering
\caption{Prompt for \textit{offline} caption pre-extraction.}
\resizebox{\columnwidth}{!}{
\begin{tabular}{p{10cm}}
\toprule
\textbf{Input Prompt} \\
\midrule
You are a professional Monster Hunter player. You are playing `Monster Hunter: World'.\\
You will receive consecutive video frames showing the battle screen as visual information for \{\textcolor{red}{entity}\}.\\
Make a `Description' for the battle screen as your `Response'. `Description' should include monster's limb and body movements, mouth, surrounding and others details.\\
Note that you should not output any information other than your `Response'.\\
Now, start to complete your task.\\
Your `Response':\\
\bottomrule
\end{tabular}
}
\label{off}
\vspace*{-4mm}
\end{table}

\begin{table}[t]
\centering
\caption{Prompt for accuracy calculation using GPT-4o as a judge.}
\resizebox{\columnwidth}{!}{
\begin{tabular}{p{10cm}}
\toprule
\textbf{Input Prompt} \\
\midrule
You are a professional Monster Hunter player. You are playing `Monster Hunter: World'.\\
Here is a `Question' need to be answered: \{\textcolor{red}{question}\}. \\
There are also two answers for this `Question': \\
Answers 1: \{\textcolor{red}{answer gt}\}. \\
Answers 2: \{\textcolor{red}{answer pred}\}. \\
Your `Response' is to decide whether the content of these two answers are similar. \\
If similar directly output `Yes'. \\
If not similar directly output `No'. \\
Note that you may ignore the format difference. \\
Ignore the difference of monster name before word, e.g., Zinogre Leap Attack and Leap Attack are with same meaning. \\
    
Here are some samples for decide similarity: \\
Sample 1: \\
`Question': Tell me what is the specific name of attack action that Zinogre is performing? \\
``Answer 1": Static Charge \\
``Answer 2": Thunder Charge B \\
``Response": No \\
Sample 2: \\
`Question': Start with counterattack, Zinogre released the attack action shown in the input battle screen. Tell me what is the next attack action? \\
``Answer 1": Zinogre Back Slam \\
``Answer 2": Back Slam \\
``Response": Yes \\
Sample 3: \\
`Question': What attack action Brachydios is unleashing? \\
``Answer 1": Brachydios is unleashing the Brachydios Ground Slime Explosion attack \\
``Answer 2": Ground Slime Explosion \\
``Response": Yes \\
 
Note that you should not output any information other than your `Response'. \\
Now, start to complete your task. \\
Your `Response': \\
\bottomrule
\end{tabular}
}
\label{compare}
\vspace*{-4mm}
\end{table}

\begin{table}[t]
\centering
\caption{Prompt for similarity calculation between generated and human-made caption using GPT-4o as a judge.}
\resizebox{\columnwidth}{!}{
\begin{tabular}{p{10cm}}
\toprule
\textbf{Input Prompt} \\
\midrule
You are a professional Monster Hunter player. You are playing `Monster Hunter: World'.\\
Here are two text description of a monster attack action.\\
Your `Response' is to decide whether the content of these two text descriptions are similar.\\
Your should focus on the details of movement and some key information that can help you to discriminate the action.\\
If similar directly output `Yes'.\\
If not similar directly output `No'.\\
    
The First description is \{\textcolor{red}{truth}\}.\\
The Second description is \{\textcolor{red}{generated}\}.\\

Note that you should not output any information other than your `Response'.\\
Now, start to complete your task.\\
Your `Response':\\
\bottomrule
\end{tabular}
}
\label{judge}
\vspace*{-4mm}
\end{table}

\subsection{Knowledge Consistency Calculation}
As defined in the main paper, the model's final output is a retrieved subgraph, denoted as $\hat{\mathcal{I}}$. We consider each path from the root entity to a leaf entity as a unique knowledge instance and represent the set of such paths as $\hat{\mathcal{L}}$. The knowledge consistency is computed between $\hat{\mathcal{L}}$ and the ground-truth knowledge paths $\mathcal{L}$ using a one-to-one matching approach.

The recall and precision of retrieved knowledge paths are defined as follows:

\begin{equation}
\text{Recall} = \frac{|\hat{\mathcal{L}} \cap \mathcal{L}|}{|\mathcal{L}|}
\end{equation}

\begin{equation}
\text{Precision} = \frac{|\hat{\mathcal{L}} \cap \mathcal{L}|}{|\hat{\mathcal{L}}|}
\end{equation}

where $\hat{\mathcal{L}} \cap \mathcal{L}$ represents the set of correctly retrieved knowledge paths. Recall measures the proportion of ground-truth knowledge paths successfully retrieved by the model, while precision measures the proportion of retrieved paths that are correct.

\subsection{Human Evaluation of GPT-4o as a Judge}\label{name}
Tables \ref{compare} and \ref{judge} present the templates for using GPT-4o as a judge \cite{zheng2023judging} to assess result accuracy (\textit{Acc.}) and caption similarity (\textit{Sim.}). For accuracy evaluation, we prompt GPT-4o to compare the similarity between the ground-truth answer \{answer gt\} and the generated answer \{answer pred\}. Additionally, we provide three few-shot examples as references for the model.

For caption similarity assessment, GPT-4o directly compares the human-written caption \{truth\} with the model-generated caption \{generated\}. To further evaluate GPT-4o's judging performance, we conducted a human experiment. As shown in Table \ref{fig:human_eval}, two knowledgeable players independently evaluated 200 randomly selected samples from GPT-4o's judgments across all experiments for each model. A judgment was considered correct if both evaluators agreed. Our findings indicate that while there are some variations across models, GPT-4o demonstrates a high overall accuracy in judgment (0.926). Although caption similarity scoring is lower, it remains sufficiently high for such a subjective task. Overall, the results show that using GPT-4o as a judge is with high feasibility.

\begin{table}[t]
\caption{Human evaluation for GPT-4o judgment accuracy. Each model's generation for answer and caption is evaluated by 200 randomly select samples through two knowledgeable players.}
\centering
\resizebox{0.95\columnwidth}{!}{
\begin{tabular}{lcc}
\toprule
 & Answer Accuracy & Caption Similarity \\
\midrule
GPT-4o \cite{achiam2023gpt} & 0.925& 0.865 \\ 
Claude 3.7 Sonnet \cite{TheC3} & 0.900 & 0.840 \\
Ovis2-16B \cite{lu2024ovis} & 0.955 &  0.810 \\
\midrule
average & 0.926 & 0.838 \\
\bottomrule
\end{tabular}}
\label{fig:human_eval}
\end{table}

\subsection{Additional Experiments for MH Benchmark}
In Table \ref{question_abaltion}, we present the impact of incorporating the monster's name (Name) and additional information (Extra) as part of the input question $q$. The metric \textit{Top.} represents the accuracy of the model in selecting the correct topic entity as the retrieval root. We observe that removing the monster's name leads to a significant performance drop due to incorrect root entity selection (low \textit{Top.}).

Additional information refers to contextual hints, such as a monster being angry, which players can infer from the game's text. These details are generally too subtle to be captured from images by MLLMs. Removing only the additional information also results in an obvious performance drop, indicating that such visually independent cues are essential for the model to generate the correct answer. One interesting observation is that with additional information \textit{Top.} can be improved than no Name and Extra setting.

\begin{table}[!t]
    \centering
    \caption{Impact of having monster name and Extra information in question. \ding{52} means having such information.}
    \resizebox{0.75\columnwidth}{!}{
    \begin{tabular}{llcccc}
        \toprule
         & & \multicolumn{3}{c}{Unaided-Online} &\\
        \cmidrule(lr){3-5} 
        Name & Extra & \textit{Acc.} & \textit{Pre.}  & \textit{Rec.} & \textit{Top.}\\ 
        \midrule
         &  &.2731  & .1251 & .2413 & .5210\\
               &\ding{52} & .3781 & .2080 &.4434 & .7365 \\
        \ding{52} &  & .4075  & .2120 & .4636& 1\\
        \ding{52} &\ding{52} & .5105 & .2756 & .5625 & 1\\
        \bottomrule
    \end{tabular}
    }
    \label{question_abaltion}
\end{table}

Table~\ref{effeceint} reports average retrieval time (in seconds), number of agent calls (n rounds), and per-call response time in the format of mean ± std. Experiments were conducted using GPT-4o and Gemini 2.0 Flash via API, and InternVL2.5 on a local GPU server with 2 A6000. The results reveal efficiency as a limitation of the current agent pipeline. More results will be included.

\begin{table}[!t]
    \caption{Computational cost per sample in average.}
    \vspace{-0.1in}
    \centering
    \resizebox{1\columnwidth}{!}{
    \begin{tabular}{lccc}
        \toprule
        Models & Retrieval Time (s) & Agent Call (n) & Response Time (s) \\ 
        \midrule
        GPT-4o  & 92.46 ± 68.93 & 7.20 ± 5.01  & 10.92 ± 9.70   \\
        Gemeni 2.0 Flash & 17.15 ± 10.92 & 11.04 ± 9.42 & 1.12 ± 0.91   \\
        InternVL & 57.06 ± 41.78 & 9.32 ± 3.58 & 7.33 ± 6.95  \\
        \bottomrule
    \end{tabular}
    }
    \label{effeceint}
    \vspace{-0.1in}
\end{table}

We also perform ablation studies to assess the impact of using cross-models for two key agents: \textit{Summarizer} (knowledge utilization) and \textit{Validation} (knowledge retrieval), keeping other agents fixed. Table~\ref{ablation} shows \textit{Acc.} results across GPT-4o, Claude 3.7, and InternVL2.5, with diagonal values representing the results of original single-model pipeline. \textit{Summarizer} replacement yields little change between GPT-4o and Claude, indicating that performance gains stem more from retrieved knowledge quality than summarization strength (InternVL’s column with better knowledge, the improvement is more evident than that in the rows, showing in green). In contrast, using a weaker model (InternVL) for \textit{Validation} causes a sharp performance drop (in red), underscoring the importance of this role. Yet, upgrading only the \textit{Validation} agent in InternVL brings limited benefit, suggesting other retrieval-stage agents affect a lot.

\begin{table}[!t]
    \caption{Ablation for cross-models agent pipeline.}
    \vspace{-0.1in}
    \centering
    \resizebox{1\columnwidth}{!}{
    \begin{tabular}{lcccccc}
        \toprule
         &  \multicolumn{3}{c}{Replace \textit{Summarizer} With} & \multicolumn{3}{c}{Replace \textit{Validation} With}\\
        \cmidrule(lr){2-4} \cmidrule(lr){5-7}
        Models & GPT-4o & Claude  & InternVL & GPT-4o & Claude  & InternVL\\ 
        \midrule
        GPT-4o & .5105 & .4994  & .4864\cellcolor{green!30}   & .5105 & .4716 & .4128\cellcolor{red!37} \\
        Claude & .4510  & .4338 & .4086\cellcolor{green!22} & .5052 & .4338 & .3676\cellcolor{red!22} \\
        InternVL & .3876\cellcolor{green!12} & .3624\cellcolor{green!9} & .3080   & .3413 & .3225 & .3080  \\
        \bottomrule
    \end{tabular}
    }
    \label{ablation}
    \vspace{-0.2in}
\end{table}

\subsection{More Result Samples}
This section presents some randomly selected examples of generated answers via various models. 

\begin{figure}[t]
\centering
\includegraphics[width=0.8\columnwidth]{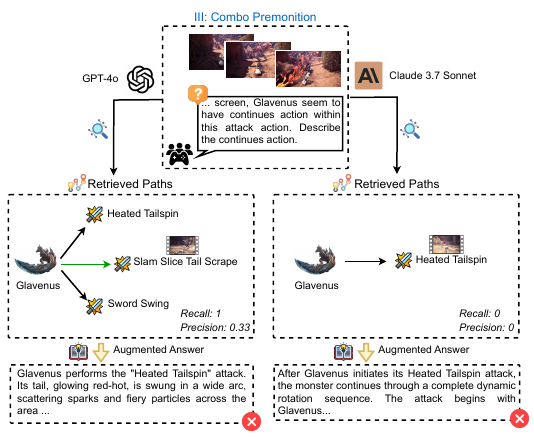}
\caption{A sample for ``Glavenus" continues attack recognition.}
\label{fig:spp1}
\end{figure}

Figure \ref{fig:spp1} shows a sample for ``Glavenus" continues attack action recognition. Both GPT-4o and Claude 3.7 output wrong answer, although GPT-4o catch the path towards true knowledge. Show models lack the ability to comprehend the knowledge.

\begin{figure}[t]
\centering
\includegraphics[width=0.8\columnwidth]{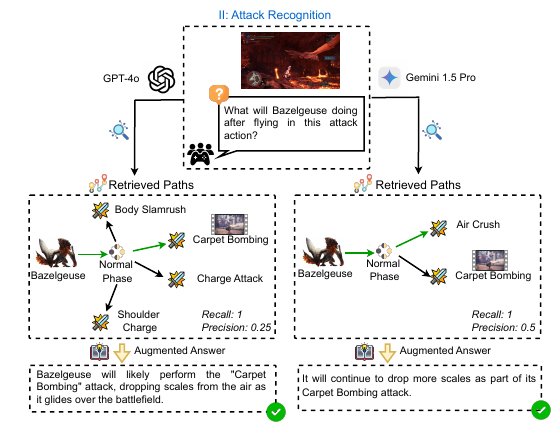}
\caption{A sample for ``Bazelgeuse" attack action recognition.}
\label{fig:spp2}
\end{figure}

Figure \ref{fig:spp2} shows a sample for ``Bazelgeuse" attack action recognition. Although some difference in response, both GPT-4o and Gemini 1.5 Pro generate correct answer. GPT-4o find more paths as its knowledge augmentation. 

\begin{figure}[t]
\centering
\includegraphics[width=0.8\columnwidth]{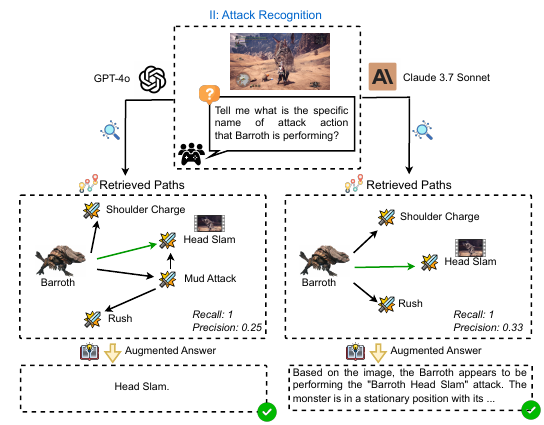}
\caption{A sample for ``Barroth" attack action recognition.}
\label{fig:spp3}
\end{figure}

Figure \ref{fig:spp3} shows a sample for ``Barroth" attack action recognition. Both GPT-4o and Claude 3.7 generate the correct answer, however, GPT-4o's answer is more clear, showing better instruct following ability.

\end{document}